\documentclass[11pt,a4paper]{article}
\usepackage[hyperref]{acl2021}
\usepackage{times}
\usepackage{latexsym}

\usepackage{caption}
\usepackage{subcaption}
\usepackage{graphicx}
\usepackage{booktabs}

\usepackage{amsmath}
\usepackage{adjustbox}

\usepackage{multirow}
\usepackage{longtable}

\usepackage[utf8]{inputenc}
\usepackage{microtype}

\usepackage{array}
\usepackage{xspace}

\usepackage{xcolor,colortbl}
\usepackage{tikz}
\usepackage{collcell}
\usepackage{xstring}
\usepackage{textcomp}

\usepackage{hyperref}
\usepackage{xfrac}
\usepackage{afterpage}

\DeclareSymbolFont{extraup}{U}{zavm}{m}{n}
\DeclareMathSymbol{\varheart}{\mathalpha}{extraup}{86}
\DeclareMathSymbol{\vardiamond}{\mathalpha}{extraup}{87}

\aclfinalcopy 
\def\aclpaperid{3753}

\newcolumntype{P}[1]{>{\centering\arraybackslash}p{#1}}

\newcommand{\custo}{\bf \small \fontfamily{qcs}\selectfont}

\newcommand{\confacl}{\mbox{{\custo ACL}}\xspace}
\newcommand{\confcl}{\mbox{{\custo CL}}\xspace}
\newcommand{\confcoling}{\mbox{{\custo COLING}}\xspace}

\newcommand{\confeacl}{\mbox{{\custo EACL}}\xspace}
\newcommand{\confemnlp}{\mbox{{\custo EMNLP}}\xspace}
\newcommand{\conflrec}{\mbox{{\custo LREC}}\xspace}
\newcommand{\confnaacl}{\mbox{{\custo NAACL}}\xspace}

\newcommand{\conftacl}{\mbox{{\custo TACL}}\xspace}
\newcommand{\confws}{\mbox{{\custo WS}}\xspace}

\newcommand{\otherconf}[1]{\mbox{{\custo #1}}\xspace}

\newcommand{\aspects}{{\em aspects}\xspace}
\newcommand{\topics}{{\em topics}\xspace}

\title{Use of Formal Ethical Reviews in NLP Literature:\\ Historical Trends and Current Practices}

\author{ \textbf{Sebastin Santy$^{\spadesuit}$} \quad \textbf{Anku Rani$^{\heartsuit}$} \quad \textbf{Monojit Choudhury$^{\spadesuit}$} \\
$^{\spadesuit}$ Microsoft Research, Bangalore, India \\
$^{\heartsuit}$ Plaksha University, Mohali, India \\
$^{\spadesuit}$\{\href{mailto:t-sesan@microsoft.com}{t-sesan}, \href{mailto:monojitc@microsoft.com}{monojitc}\}{@microsoft.com}, $^{\heartsuit}$\href{mailto:anku.rani@plaksha.org}{anku.rani}@plaksha.org
}

\date{}

\begin{document}
\maketitle
\begin{abstract}
Ethical aspects of research in language technologies have received much attention recently. It is a standard practice to get a study involving human subjects reviewed and approved by a professional ethics committee/board of the institution. How commonly do we see mention of ethical approvals in NLP research? What types of research or aspects of studies are usually subject to such reviews? With the rising concerns and discourse around the ethics of NLP, do we also observe a rise in formal ethical reviews of NLP studies? And, if so, would this imply that there is a heightened awareness of ethical issues that was previously lacking?
We aim to address these questions by conducting a detailed quantitative and qualitative analysis of the ACL Anthology, as well as comparing the trends in our field to those of other related disciplines, such as cognitive science, machine learning, data mining, and systems.

\end{abstract}

\section{Introduction}

With the rapid advances in the field of Natural Language Processing (NLP), language technologies are getting woven into the daily fabric of our lives and the society.  Since ``language is a portal of emotions, a proxy of human behavior, and a strong signal of individual characteristics" \cite{hovy2016social}, large-scale deployment of language technology has potential risks that require early detection and mitigation. Naturally, there have been several discussions about the potential harms and ethical issues concerning NLP \cite{hovy2016social, conway2016social}. They have mostly revolved around building or deploying systems in sensitive areas such as hate speech \cite{sap2019risk}, social media \cite{benton2017ethical}, clinical NLP and mental health \cite{vsuster2017short, mikal2016ethical} and use of sensitive or personal information \cite{larson2017gender}. While building NLP systems, there are also ethical risks associated with involvement of human subjects through user studies or data collection activities \cite{shmueli2021beyond}. 

\begin{figure}[t]
    \centering
    \includegraphics[width=\linewidth]{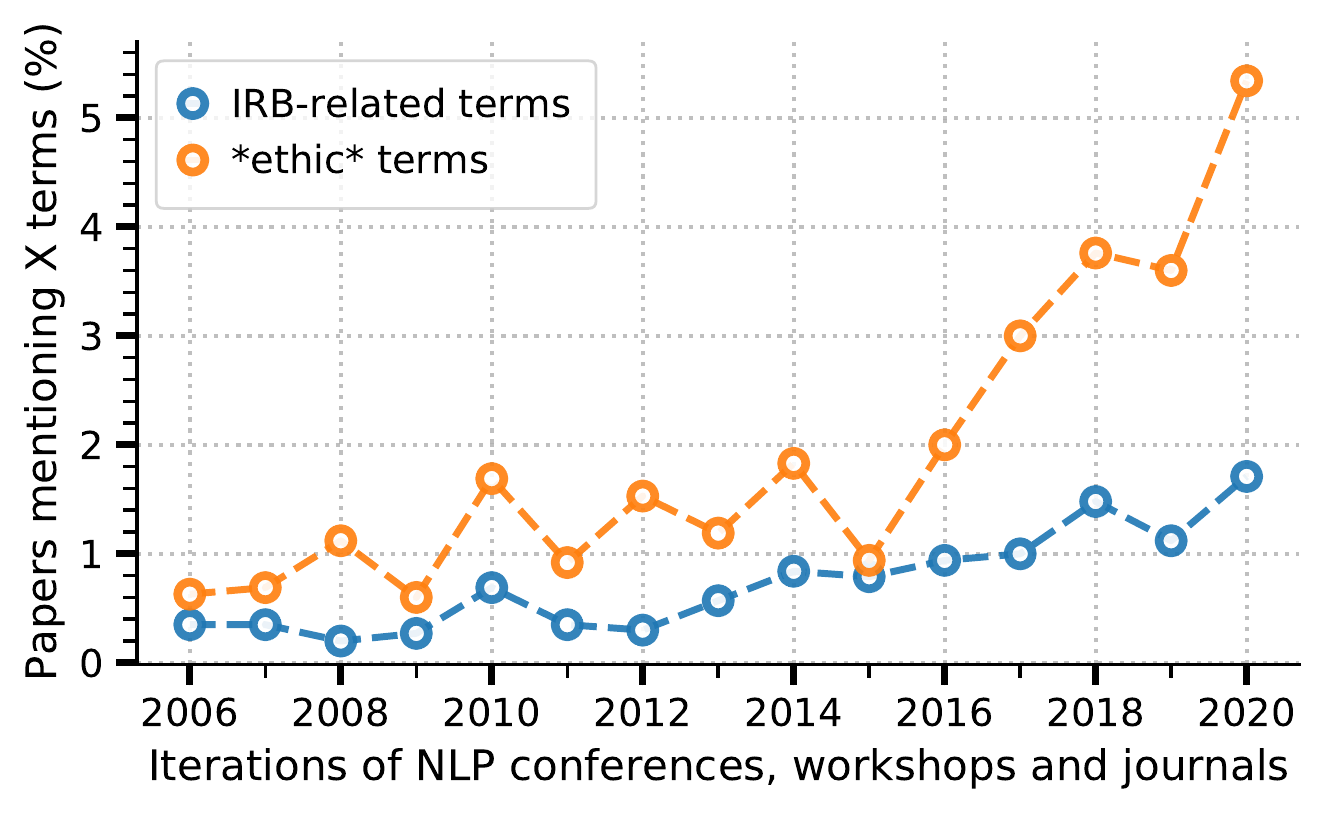}
    \caption{\small Percentage(\%) of papers mentioning \underline{IRB-related} and \underline{*ethic*} terms in NLP conferences, journals, and workshops from 2006 to 2020.}
    \label{fig:irbacl}
\end{figure}

The awareness of the dangers of the existing and new NLP applications has led to the curation of several ethical guidelines and frameworks. Undergirded by lessons from the past, these guidelines and frameworks help researchers consider and contextualize critical ethical concerns. Most of the ethical issues in NLP are rooted in the data being used for research. \citet{couillault2014evaluating} is one of the first works to explore the ethics of data collection and evaluation in NLP.  Several other works have proposed best practices for dealing with ethical implications of NLP research and deployment \cite{prabhumoye2019principled, leidner2017ethical, bender2018data, schnoebelen2017goal}. There is now an increased awareness around this topic with a number of workshops and tutorials on ethics at NLP conferences \cite{tsvetkov2018socially, ws-2017-acl, ws-2018-acl}. Such discussions have resulted in a number of reforms at NLP conferences. NLP conferences now have new track called {\em Ethics in NLP}. Furthermore, several ML and NLP conferences such as \otherconf{NeurIPS} 2020, \confnaacl 2021 and \confacl 2021\footnote{\url{https://2021.aclweb.org/ethics/Ethics-FAQ/}} now recommend the inclusion of broader impact statement in their papers, which allows for authors to introspect and be mindful of the ethical implications their research poses.

Although an NLP researcher might be individually committed towards ethical research practices, they may not have the expertise in ethical and legal issues to gauge the potential risks of a technology or a dataset that they are building. The practice of getting the research/study approved by an ethical review board (aka Institutional Review Board or IRB)\footnote{In this paper, we use IRB as a generic term to refer to such review boards which are known by slightly different terms in different geographies.} instituted by the organization is thus critical in early defusal of the potential harms. The two primary functions of IRB are (i) to protect the rights and welfare of human research subjects, and (ii) to support and facilitate the conduct of valuable research \cite{bankert2006institutional, klitzman2012institutional, byerly2009working}. Traditionally, IRB has been a long-standing norm in biomedical research due to its overt exposure to human subjects. However, with computing research pervading human lives, IRBs have started covering computing research as well \cite{buchanan2009internet, vitak2017ethics}. With regards to NLP, most of the data collection and annotation processes as well as user studies come under the purview of these boards. These are particularly necessary if they cover sensitive topics such as mental health or hate speech which can affect the human subjects involved in data collection or the users of the system. 

How frequently do NLP researchers take IRB approvals for their studies? What aspects of NLP research or which topics of study are typically considered for IRB approvals? What are the historical and current trends, and what can we say about the awareness of the NLP research community around ethical issues? In this paper, we try to answer these questions through a quantitative and qualitative analysis of papers from the ACL anthology that seek IRB approvals. According to our findings, IRB approvals were almost non-existent in the NLP literature until 2006, but there has been a steady increase since 2016. We also study the distribution of IRB approvals by country and industry/academia affiliation, as well as compare the recent trends in NLP conferences to that of various prominent conferences ranging from machine learning and data mining to human-computer interaction and systems. One of the key findings of this study is that IRB permission was mostly sought for either data collection or annotation studies, but hardly ever for data re-purposing or system design/deployment - a void that we think the NLP community should be conscious about.

\section{Method}
To determine the trends of IRB approvals in NLP research, we resort to searching for IRB- and ethics-related terms in research papers. We obtain the papers (PDFs) for major NLP conferences, journals and workshops [\confacl, \confcoling, \confeacl \confemnlp, \conflrec, \confnaacl, \confcl, \conftacl, and \mbox{{\custo WS}}] from the ACL Anthology (curated by \citet{joshi-etal-2020-state}). For a comparative analysis, we also collect papers from other related conferences [\otherconf{CogSci}, \otherconf{InterSpeech}, \otherconf{NeurIPS}, \otherconf{CVPR}, \otherconf{ICWSM}, \otherconf{CHI}, \mbox{{\custo COMPASS}}] for the years 2019 and 2020, during which there was considerably more discussion around ethics of computing research.

In order to retrieve papers that seek IRB approvals, we search for the following keywords which cover the phrases used for IRB in countries that are frequently represented at these conferences: \underline{review board}, \underline{ethics board}, \underline{\smash{ethics panel}}, \underline{ethics committee}, \underline{consent form}, and \underline{IRB}.\footnote{Collectively referred to as \underline{IRB-related} terms from hereon.} To further compare and calibrate, we also search for papers that contain the wildcard string \underline{*ethic*}, which brings up a broader set of papers that may discuss ethical repercussions of their work, even if any approval is not explicitly sought or mentioned. To assist with a robust search over this textual data, we use the Allen AI SPIKE interface\footnote{URL when accessed: \url{https://spike.staging.apps.allenai.org/datasets/acl/search}} \cite{taub-tabib-etal-2020-interactive,shlain-etal-2020-syntactic} and use pdfgrep\footnote{Command-line tool: \url{https://pdfgrep.org/}} to cross-check our results.

\underline{IRB-related} term search yielded $210$ papers from the ACL anthology (till 2019), which were then manually checked for precision and annotated for \aspects (see Figure~\ref{fig:taxonomy})  and \topics (e.g., hate speech, social media, mental health, etc.) of the research for which IRB permission was sought. Through our manual curation, we found that $94.17\%$ of these papers actually took the approval for their research study thus showing that our search is precise in capturing the terms. The remaining papers were mostly ethical frameworks and recommendations (e.g., \citet{hovy2016social, bender2018data}) which merely mentioned the need for seeking IRB approvals in NLP research.

\section{Findings}
\subsection{How many papers seek IRB approvals?}
\begin{figure}[h]
    \centering
    \includegraphics[width=\linewidth]{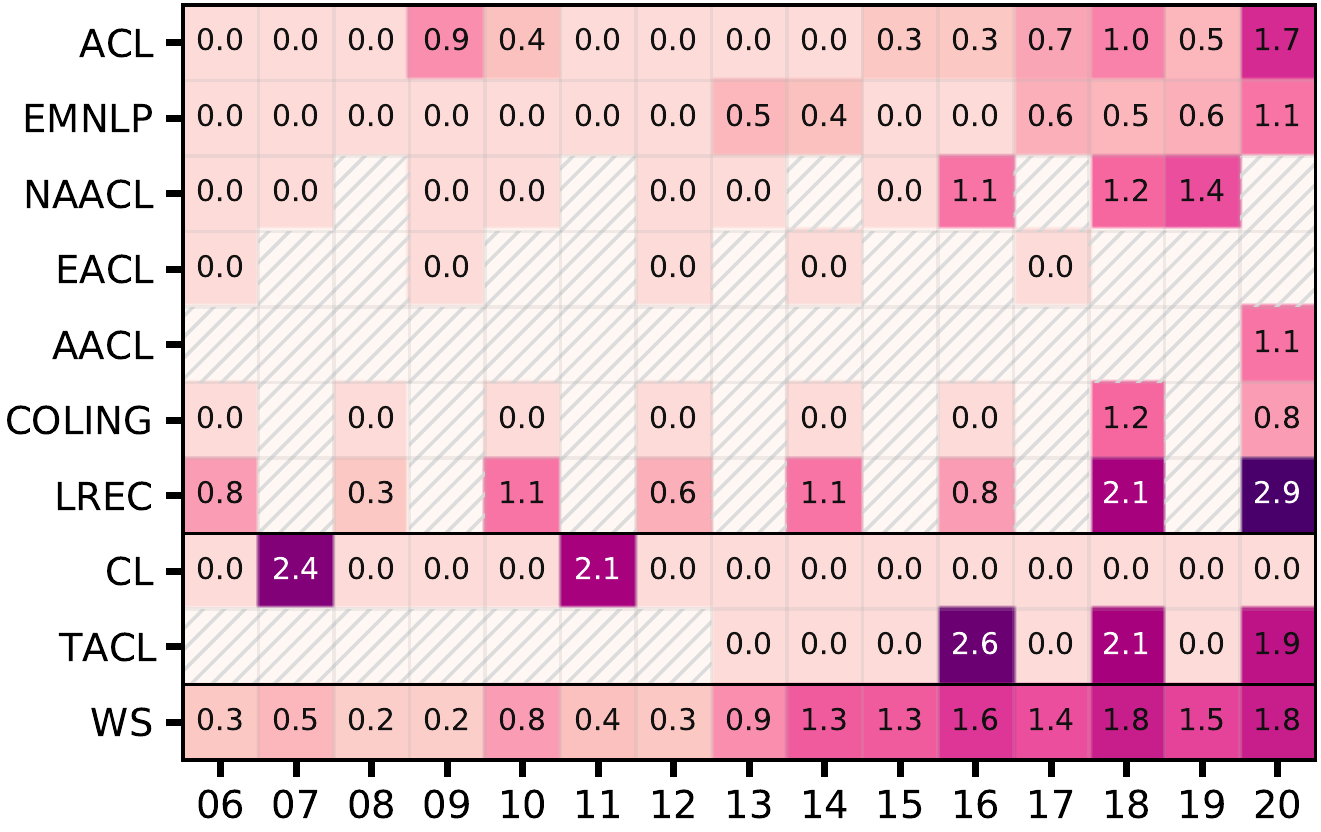}
    \caption{\small The percentage (\%) of papers in NLP conferences, journals and workshops over the past 15 years that mention \underline{IRB-related} terms. The intensity of color is proportional to the \% values. The boxes with a gray hatch reflect the years when that particular conference was not held.}
    \label{fig:irboveryears}
\end{figure}
Figure \ref{fig:irboveryears} shows the percentage(\%) of papers in each NLP conference iteration that mention \underline{IRB-related} terms. It is immediately obvious that for almost all the conferences only a minuscule number of papers mention IRB approvals. However, it is heartening to see that the number of mentions is increasing in recent years. \conflrec and \confws particularly stand out among the other conferences for having at least some mentions of IRB approvals in every iteration. For \conflrec, it is understandable, since the theme of the conference revolves around data and resource generation. In the case of \confws, IRB mentions are consistently increasing over the years. We observed that this is mostly due to the diverse nature of workshops some of which are on resource generation \cite{popescu2019proceedings} or cover sensitive topics \cite{niederhoffer2019proceedings, yannakoudakis2019proceedings}. Journals such as \confcl and \conftacl have very few papers in each iteration, so even one IRB mention appears to be a lot.  It should be noted that there is a possibility that a research study has obtained IRB approval but has not disclosed it in their paper. However, based on the authors' experience (and anecdotes from personal conversations), it is highly unlikely that anyone who has been through the IRB approval process will fail to mention it.

\subsection{What kinds of research seek IRB approvals?}
\begin{figure}[h]
    \centering
    \includegraphics[width=0.9\linewidth]{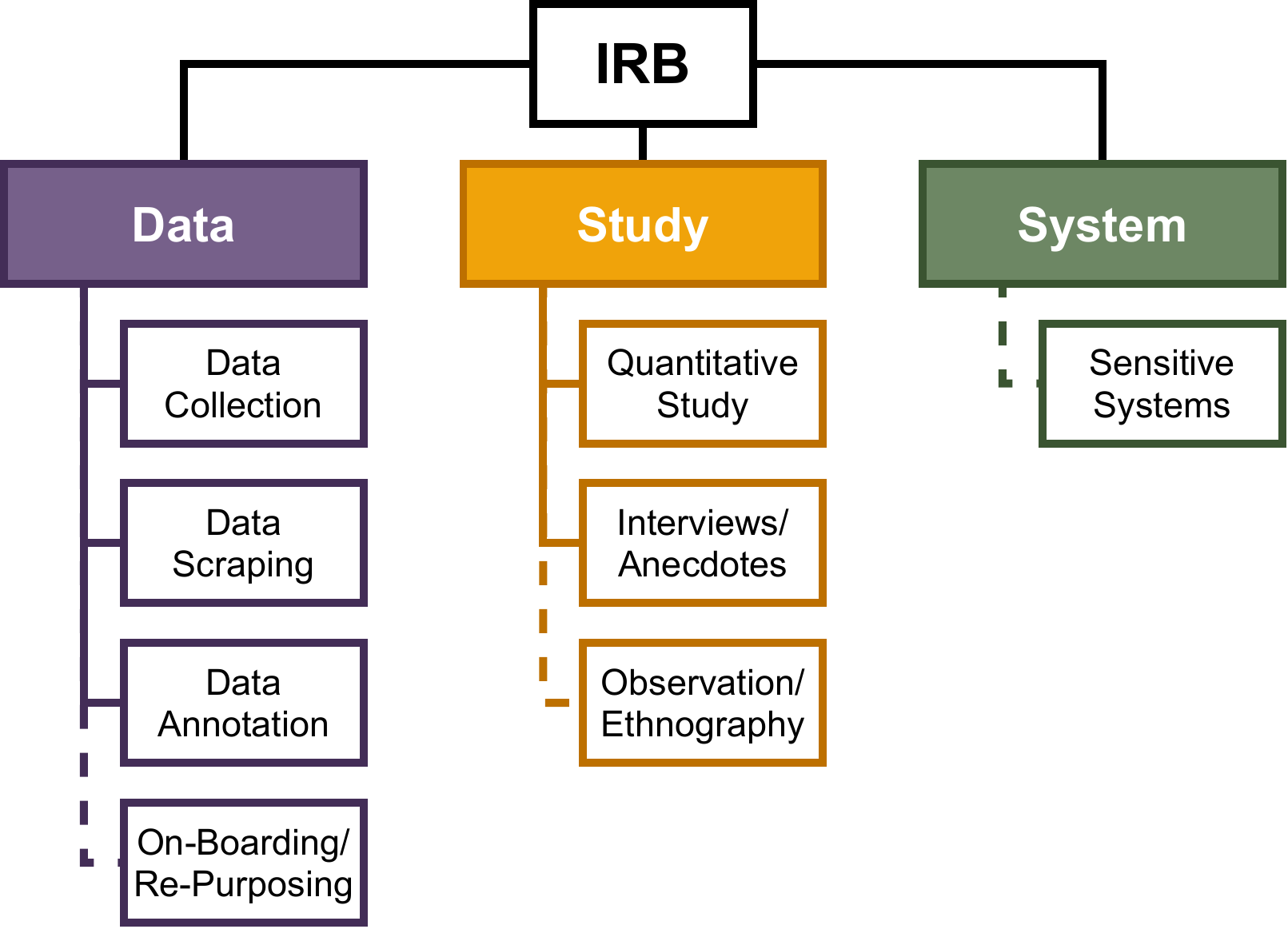}
    \caption{\small Different \aspects of research for which IRB approvals were sought in the papers that we manually analyzed.}
    \label{fig:taxonomy}
\end{figure}
\begin{figure*}[t]
    \centering
    \includegraphics[width=\linewidth]{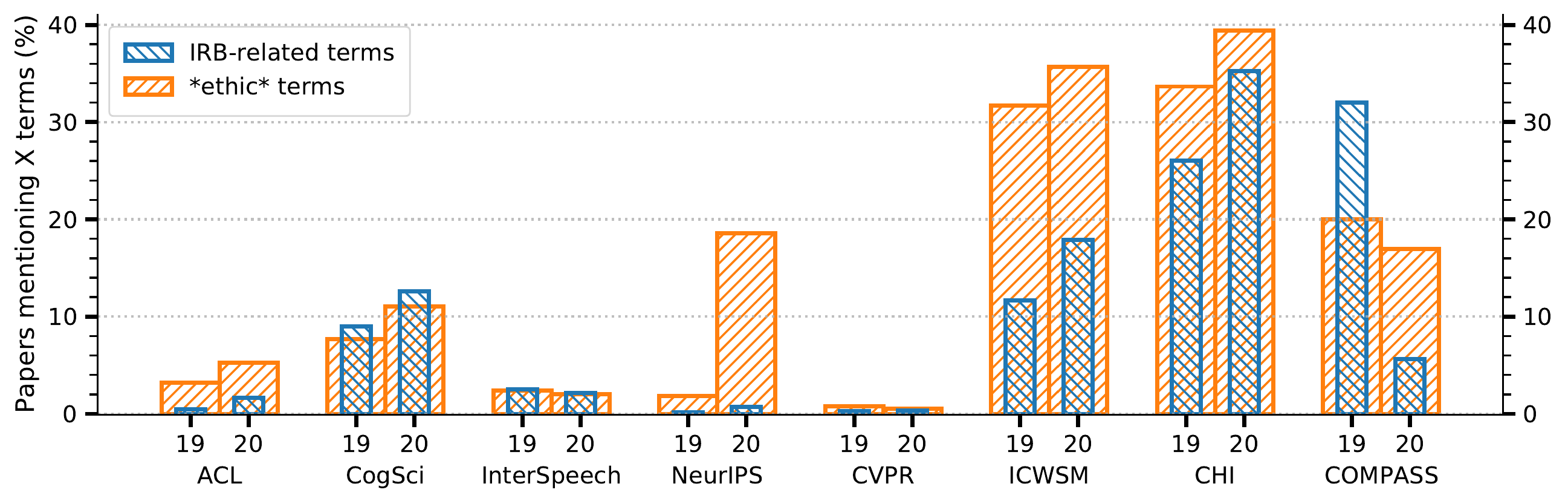}
    \caption{\small Percentage(\%) of papers mentioning \underline{IRB-related} and \underline{*ethic*} terms in related conferences.}
    \label{fig:irbconfs}
\end{figure*}
We manually go through each of the 58 NLP papers (which excludes \confws) to derive the \aspects and understand the context in which IRB approvals are sought and build a taxonomy of the broad topics covered (Figure \ref{fig:taxonomy}). We see that most of them ($24$ papers; $41.3\%$) take IRB approvals for collection of data which can often involve human subjects directly. It is followed by the annotation of data with $20$ papers ($34\%$) taking IRB approvals. A meager $7$ papers ($12\%$) in our set take IRB approvals for scraping data which is the automatic collection of data from web pages or social media posts without explicit consent from the users. We see that only one paper takes IRB approval for re-purposing and further annotation of data \cite{rogers2018rusentiment}. One of the core concerns of GDPR is the usage of personal data collected by media platforms for a purpose different than what the user consented to and hence such re-purposing of data should ideally undergo IRB approvals. $12$ papers ($20\%$) take approvals for conducting user studies of both qualitative (survey, interview) and quantitative nature (semantic edits). Interestingly, we see that only one paper has taken IRB approval for the whole system owing to its sensitive nature \cite{cao2018baby}.

We also look at the nature (or \topics) of the work for which the IRB approvals are taken. We observe that $48.4\%$ of papers that mention IRB have sensitive topics (such as mental health, hate speech, clinical/medical NLP), $20.3\%$ of the papers are for collection of eye movement, EEG and audio/video recordings of human subjects, and rest of them are generic data collection or user study. To further understand the trends, we look at certain tracks of \confacl 2020 which deal with sensitive topics or data collection. We notice that only $1/42$ {\em Resource and Evaluation} papers and $3/24$ {\em Computation Social Science} papers have taken IRB approvals. It is worth noting that $29/210$ papers we have annotated only used an informed consent form without explicitly mentioning whether an ethical review board was involved in the process.

\subsection{{What is the distribution of IRB approvals by country and industry/academia?}} 

\begin{table}[h]
\small
\centering
\begin{tabular}{@{}lrr@{}}
\toprule
 & \textbf{Total Papers} & 
 \textbf{Percent Papers}\\ \midrule
\textbf{Countries} &  & \\
\addlinespace[0.5em]
USA & $47/5368$ & $0.88\%$ \\
\addlinespace[0.2em]
Canada & $5/358$ & $1.40\%$ \\
\addlinespace[0.2em]
Germany & $5/850$ & $0.59\%$ \\
\addlinespace[0.2em]
UK & $5/1088$ & $0.46\%$ \\
\addlinespace[0.2em]
Netherlands & $3/226$ & $1.33\%$ \\
\addlinespace[0.2em]
Sweden & $2/100$ & $2.00\%$ \\
\addlinespace[0.2em]
South Korea & $2/151$ & $1.32\%$ \\
\addlinespace[0.2em]
China & $2/2350$ & $0.09\%$ \\
\addlinespace[0.2em]
\midrule
\textbf{Affiliation Types} &  &  \\
\addlinespace[0.5em]
University & $52/7730$ & $0.67\%$ \\
\addlinespace[0.2em]
Industry & $1/841$ & $0.12\%$ \\
\addlinespace[0.2em]
National Lab & $1/182$ & $0.55\%$ \\
\addlinespace[0.2em]
Joint/Collaboration & $11/2651$ & $0.41\%$ \\
\addlinespace[0.2em]

\bottomrule
\end{tabular}
\caption{\small Distribution of \% \underline{IRB-related} term mentions among countries and different types of affiliations for NLP conferences (excluding \conflrec and \confws) from 2012 to 2020. \footnotemark}
\label{tab:irbgeoindus}
\end{table}
\footnotetext{Country and Affiliation data obtained from \url{https://github.com/marekrei/ml_nlp_paper_data/}}

Table \ref{tab:irbgeoindus} shows the distribution of papers which mention IRB approvals along two dimensions: countries and types of institutions. As can be observed, most of the listed countries are WEIRD\footnote{Western, Educated, Industrialized, Rich and Democratic} societies. 
When it comes to the type of institution, we find that universities account for the vast majority of papers seeking IRB approvals, followed by joint collaborations. This trend can be counter-intuitive as an industry is more likely to be regulated and accountable for the ethical and legal concerns of their work. One possibility is that industries perhaps do not engage in external data collection/annotation work or conduct user studies as much as academic institutions do. Alternatively, it is possible that the data collection/annotation process is a completely independent pipeline that is not specific to the research paper in which it is used and thus is not reported.

\subsection{{How do the IRB trends in NLP research compare with those in related fields?}}

We look at the following conferences for our analysis: \otherconf{CogSci} in cognitive science, \otherconf{InterSpeech} in speech processing, \otherconf{NeurIPS} in machine learning, \otherconf{CVPR} in computer vision, \otherconf{ICWSM} in social media mining, \otherconf{CHI} in human-computer interaction, and \otherconf{COMPASS} in computing systems deployment. We specifically analyze these conferences for 2019 and 2020 iterations as there have been significant changes made in the conferences during this period in terms of reporting the ethical ramifications of their research. Figure \ref{fig:irbconfs} shows the \% of papers mentioning \underline{*ethic*} and \underline{IRB-related} terms for each conference iteration. We calculate for \underline{*ethic*} to understand how aware and concerned each field/conference is towards the ethical implications of the research they conduct.

It is not surprising that IRB mentions for \otherconf{CHI} are so high ($\sim35\%$) given that more than $65\%$ percent of CHI papers include at least one user study \cite{koeman-hcistudies}. \otherconf{ICWSM} works with datasets and systems related to web and social media analytics and hence would need to undergo IRB approvals. This is apparent in the relatively high number of IRB mentions in both 2019 and 2020. Unlike the other conferences we choose for our analysis, \otherconf{CogSci} is a non-computing conference. Linguistics work is frequently found in \otherconf{CogSci}, which often makes use of human subjects. We observe that it has the most consistent representation of both \underline{*ethic*} and \underline{IRB-related} term mentions among the years. As previously discussed, one of the concerns is that sensitive systems are seldom taking IRB approvals. On the contrary, we notice that \otherconf{COMPASS}, a conference largely focused on deploying computing systems, is prevalent in taking IRB approvals. \otherconf{InterSpeech} and \otherconf{CVPR} have significantly fewer papers with IRB mentions ($< 0.35\%$ and $<2.5\%$, respectively) and the trends have hardly changed over the years, despite the fact that they conduct research with speech, multimodal, and vision data that may have been collected from human subjects. Among these, there is a ray of hope for \confacl which is showing a significant positive trend in both \underline{*ethic*} and \underline{IRB} mentions without any external reinforcement, thanks to the increasing awareness in the field. \otherconf{NeurIPS}, on the other hand, has seen a meteoric rise in their \underline{*ethic*} mentions, which, on manual inspection reveals, is due to their mandatory inclusion of broader impact statements. There has also been a slight increase in their IRB mentions, which could be attributed to this, indicating that broader impact statements might help researchers be more cautious when proposing their research to the larger community. This quantitative testimony from \otherconf{NeurIPS} shows that \confacl and other *CL conferences are moving in the right direction with their inclusion of stringent ethics reviews for their papers.

\section{Way Forward}
In this paper, we conduct a survey of IRB approvals in NLP research. The two key observations we make are as follows. First, very few papers ($<0.8$\% of all papers published) since 2006 have sought an IRB approval; though we do observe a rise in numbers ($<1.3$\% of all papers published) since 2016. This is much smaller compared to the numbers we observe for other conferences such as \otherconf{CogSci}, \otherconf{CHI}, \otherconf{ICWSM} or \otherconf{COMPASS}.
Second, the majority of the IRB approvals were obtained for data collection or annotation that directly involved users, with only a few studies seeking approvals for data scraping or re-purposing. Such approvals are even more scarce for sensitive systems where we seldom see any paper taking IRB approvals solely for the system. The number of papers creating new datasets is expected to be greater than $1\%$ of all NLP papers\footnote{As a crude statistics, consider the fact that the number of accepted papers in the {\em Resource and Evaluation} track in \confacl 2020 was $5.4\%$, whereas only $1.7\%$ of all the papers in that year sought for IRB approvals.}; the number of papers that re-purpose an existing dataset is expected to be even greater than this. Therefore, clearly not all papers creating datasets, and almost no paper re-purposing datasets take approvals from IRB. As such, re-purposing data collected from human subjects without their explicit consent on how the data will be re-used is potentially dangerous and may even have legal repercussions. Furthermore, with the exception of a couple of papers, to date, there is no practice or trend of taking IRB approval for designing, developing, and deploying systems. This is in stark contrast to the practice in other related fields/conferences such as \otherconf{COMPASS}. Much of the harm caused by a system could actually come from its design or style of training or deployment, rather than the underlying datasets. 

We see that the broad impact statements have helped conferences such as \otherconf{NeurIPS} which were traditionally oblivious to ethical issues \cite{nanayakkara2021unpacking}. We believe that, in a similar way, the impact statements introduced in \confnaacl\footnote{Report: \url{ https://2021.naacl.org/blog/ethics-review-process-report-back/}} and \confacl 2021, with specific clauses for seeking IRB, will be highly beneficial in limiting the aforementioned potential risks by increasing the awareness amongst researchers of broader ethical repercussions of their research. It will be interesting to conduct a similar study a few years down the line and contrast with the findings of the current study.

\section*{Acknowledgements}
We thank Prof. Emily M. Bender (UW) for providing useful insights and inspiration at the early stages of this work.  We also thank Dr. Mary L. Gray and Ms. Anu Kannepalli (Microsoft) for their many insightful inputs, Prof. Yoav Goldberg (Allen AI) for setting up the  SPIKE instance for us, and Prof. Katharina Reinecke (UW) and Dr. Prasanta Bhattacharya (A*STAR) for the broad discussions around IRBs. We also thank anonymous reviewers from ACL and NLP4PI workshop for their thoughtful suggestions and feedback.

\bibliographystyle{acl_natbib}
\bibliography{acl2021}

\clearpage

\end{document}